\documentclass[a4paper,conference]{IEEEtran}
\ifCLASSINFOpdf
\else
\fi

\usepackage{times}
\usepackage{epsfig}
\usepackage{graphicx}
\usepackage{amsmath}
\usepackage{amssymb}

\begin{document}
%
\title{A Joint Representation Learning and Feature Modeling Approach for One-class Recognition}

\author{\IEEEauthorblockN{Pramuditha Perera and Vishal M. Patel}
\IEEEauthorblockA{Department of Electrical and Computer Engineering,
Johns Hopkins University\\
Email: pramuperera@gmail.com, vpatel36@jhu.edu}
}


%


\maketitle

\begin{abstract}
One-class recognition is traditionally approached either as a representation learning problem or a feature modelling problem. In this work, we argue that both of these approaches have their own limitations; and a more effective solution can be obtained by combining the two. The proposed approach is based on the combination of a generative framework and a one-class classification method.  First, we learn generative features using the one-class data with a generative framework. We augment the learned features with the corresponding reconstruction errors to obtain augmented features. Then, we qualitatively identify a suitable feature distribution that reduces the redundancy in the chosen classifier space. Finally, we force the augmented features to take the form of this distribution using an adversarial framework. We test the effectiveness of the proposed method on three one-class classification tasks and obtain state-of-the-art results.
\end{abstract}


%
\IEEEpeerreviewmaketitle

\section{Introduction}
Object classification traditionally assumes complete knowledge about all classes the classifier encounters during inference \cite{he15deepresidual},\cite{NIPS2012_ALEX},\cite{VGG}.  The availability of out-of-class training data allows the classification networks to learn discriminative features that separates each class from the rest. Recent classification networks have exploited this property to learn representations that result in superior classification performance on various datasets.

However, learning representations in the case where multiple-labelled data do not exist is an open research problem to date. In this work, we consider the extreme case where the knowledge of the classifier is limited to only a single class. In this scenario, given the training samples from a class, the classifier is expected to reject samples from any outside class. This problem is known as one-class recognition in the machine learning literature \cite{dsvdd},\cite{GPND},\cite{AND}. In practice, One-class classification has application in cases where  out-of-class examples are unavailable such as in anomaly image detection, black-box adversarial attack detection and face anti-spoofing etc.

\begin{figure}[t]
	\centering
	\includegraphics[width=0.6\linewidth]{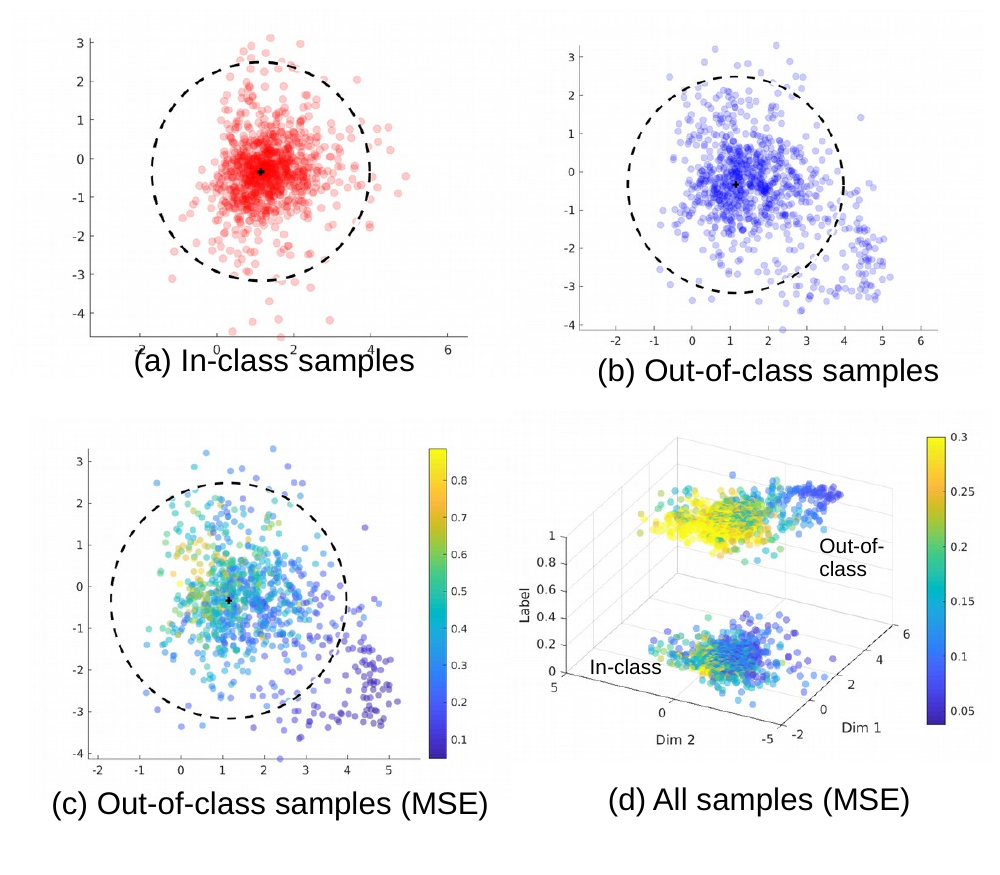}\hskip30pt
\vskip-20pt	\caption{Limitations of one-class learning strategies. Feature representation obtained for the MNIST digit 1 with 2-dimensional auto-encoder. (a) Projection of the in-class data samples on to the latent space and the obtained SVDD decision boundary (for $c=0.1$). (b) Projection of the out-of-class data on to the latent space. Some out-of-class points get projected inside the learned boundary resulting in false positive errors.  (c) Plot of the reconstruction errors of all samples. In-class and out-of-class samples are plotted on  top and bottom planes, respectively. Some out-of-class samples produce  low reconstruction errors.}
	\label{fig:issues}
\end{figure}

Historically, feature learning and classifier learning of multiple-class recognition pipeline have been treated as separate tasks \cite{Bishop:2006:PRM:1162264}. However, neural networks are known to jointly learn the classifier and features \cite{NIPS2014_5347}. For example in AlexNet \cite{NIPS2012_ALEX}, which was designed for multiple class classification, the maximum argument of the final layer (fc8) yields the classifier output. It has been shown that the penultimate layer (fc7) acts as a feature \cite{girshick14CVPR}. Therefore, when training the network, both features and the classifier are learned together. However, this is not possible in one-class applications since the performance of the learned classifier cannot be assessed. This arises as a direct result of not having any out-of-class training data.

In this light, one-class recognition algorithms have commonly followed one of the following strategies:

\noindent \textbf{1. Representation learning.} An in-class representation is learned during training. During inference, if the model is able to represent an input sample satisfactorily, it is declared to be an in-class object. For example, a deep-autoencoder trained on the given object class can be used to learn such a representation. If the reconstruction error between an input object and its representation is low, it is identified as an in-class object. This approach yields in considerable separation between in-class and out-of class samples \cite{AND}.

\noindent  \textbf{2. Feature modelling. } A  pre-determined feature extractor is used to extract features from objects/images of the given class. A one-class modeling method (such as One-class SVM (OCSVM) \cite{Scholkopf:2001:ESH:1119748.1119749}, Support Vector Data Descriptor (SVDD) \cite{Tax:2004:SVD:960091.960109}, One-class Min-max Probability Machines (OCMPM) \cite{NIPS2002_2187}) is used to identify the positive space in the feature space and to map it onto a decision boundary. Both OCSVM and OCMPM try to find a decision boundary that separates the data cloud from the origin whereas SVDD finds the tightest hyper-sphere that bounds the data without making assumptions about the underlying distribution of the data. Therefore, SVDD is  a more generic solution to one-class modeling and is proven to be effective  across a number of application domains. In this paper,  we limit our discussion to SVDD.

However, both of these approaches have their own limitations. In order to illustrate these limitations, we trained an autoencoder with a latent space size of two using the MNIST digit 1. In  Figure~\ref{fig:issues}, the SVDD decision boundary obtained for digit 1 is shown in sub-plot (a). In subplots (b) and (c) positioning of out-of-class samples with respect to the learned boundary and the corresponding reconstruction errors are  shown. Finally, in (d) a comparison of the reconstruction errors for both in-class and out-of-class samples are given.

First, there exists two limitations in the feature space modeling. First, we note that the redundant space could be identified as a part of the positive space. For example, consider the SVDD classifier boundary illustrated in Figure~\ref{fig:issues}(a). There is considerable amount of the redundant \textit{white-space} in the positive space which indicates that the classifier boundary is not sufficiently tight. This occurs because the features and the classifier are learned in sequence. Secondly, we note that there  is no guarantee that out-of-class samples will not get projected inside the identified decision boundary during inference as indicated in Figure~\ref{fig:issues}(b).

On the other hand, when a representation is learned for a given class, it can be learned such that in-class samples are well represented.  Nevertheless, there is no guarantee that out-of-class samples will not be represented well in the learned space.  For example, in Figure~\ref{fig:issues}(d), the majority of in-class samples have yielded lower reconstruction error values in the range of 0 to 0.2. At the same time, we note that some of the out-of-class samples too have produced reconstruction errors in this range. These samples will result in false positive detections.

It should be noted that feature modelling fails only when the out-of-class samples get projected inside the identified positive space. However, provided that the latent space is smooth and each latent code inside the positive space corresponds to an in-class sample, these failed cases can be identified by considering the reconstruction error (as the reconstruction error for an out-of-class sample will be higher in such conditions). Therefore, we argue that feature modelling and representation learning compliment each other and a solution can be devised such that it combines the advantages of both of these approaches.  In Table~\ref{tbl:comp}  we  compare how out-of-class samples are dealt  with under different conditions in the proposed approach with existing strategies. As shown in Table~\ref{tbl:comp}, the proposed method reduces false positive detections in three scenarios out of four as compared to the existing strategies.

\begin{table}[t!] \label{tbl:comp}
	\centering
	
	\caption{Detection output for an out-of-class sample under different strategies. Proposed method reduces false positives in three possible scenarios as opposed to the existing strategies. FP: False Positive. TN : True Negative.}
	\resizebox{0.9\linewidth}{!}{
		\begin{tabular}{|l|l|l|l|l|}
			\hline
			& \multicolumn{2}{l|}{Inside + Space} & \multicolumn{2}{l|}{Outside + Space} \\ \hline
			& Low MSE          & High MSE         & Low MSE          & High MSE          \\ \hline
			Feature Modeling        & FP               & FP               & TN               & TN                \\ \hline
			Representation Learning & FP               & TN               & FP               & TN                \\ \hline
			Proposed                & FP               & TN               & TN               & TN                \\ \hline
	\end{tabular}}
\end{table}

In our work, we learn a feature-classifier pair such that all of these factors are alleviated. As a result, our method yields better performance in one-class recognition. Specifically, we make the following contributions in this paper:

\noindent 1. Starting from an input observation, we learn a deep representation targeting one class recognition using an auto-encoder. We propose to create an augmented feature by augmenting the reconstruction error of the learned network to the latent representation produced by the auto-encoder.\\
\noindent 2.  We identify a set of desirable properties a feature distribution should satisfy to reduce the redundant space in the positive space.  We ensure that the augmented feature space is smooth and satisfies aforementioned properties using an adversarial auto-encoder.\\
\noindent 3. We obtain state-of-the-art one-class recognition performance on three publicly available datasets.

\section{Related Work}
\noindent{\textbf{Classical one-class recognition}}. One-class recognition is a well established research problem in machine learning \cite{Markou03noveltydetection}, \cite{Markou:2003:NDR:959414.959416}. Earliest attempts in one-class recognition were based on either distributional modeling or quantile estimation of the given class \cite{DBLP:conf/visapp/2009-s},\cite{DBLP:conf/visapp/2009-s},\cite{roberts}. However, in practice quantile estimation techniques have proven to be more effective compared to the distributional modeling methods.  One-class Support Vector Machines (1-SVM) \cite{Scholkopf:2001:ESH:1119748.1119749} is one of the most widely employed algorithms in one-class classification. In this algorithm, the origin of the coordinate system is treated as a pseudo-negative class. A decision boundary furthest away from the origin is learned to separate the data of the given class and the origin. However, we note that this assumption does not hold true in general. Single Class Mini-max Probability Machines \cite{NIPS2002_2187} is another classical algorithm that operates on similar principles (MPM). Different from 1-SVM, 1-MPM takes into consideration the second order statistics of the data to produce a hyper-plane that separates the unknown data from the origin. On the other hand, Support Vector Data Descriptor(SVDD) \cite{Tax:2004:SVD:960091.960109} does not make any assumption about the relative spread of out-of-class samples. Instead, SVDD constructs a tight boundary that encloses samples from the given class. Since SVDD does not make any unwarranted assumptions about the problem, it is the popular choice among the traditional one-class classifiers.

\noindent{\textbf{Deep leaning and adversarial learning.}} 
Convolutional Neural Network (CNN) based solutions have recorded state of the art performances in object recognition tasks in recent years\cite{huang2017densely},\cite{he15deepresidual},\cite{NIPS2012_ALEX}. It is very common in computer vision to use pre-trained CNN models to transfer knowledge to other inference tasks \cite{pmlr-v32-donahue14},\cite{girshick14CVPR}. However, this possibility is not considered for one-class recognition as the problem formulation states the strict use of one-class data \cite{dsvdd},\cite{AND}. Instead, auto-encoders\cite{MNISTAUTO} and variants of auto-encoders (such as de-noising auto-encoders \cite{Vincent:2008:AE},\cite{Vincent:2010:SDA:1756006.1953039} and variational auto-encoders\cite{Variational_autoencoders}) are used in one-class recognition.

Generative Adversarial Networks (GANs) introduced in \cite{goodfellow2014generative} formulated generative modeling as a two-player adversarial game. A standard GAN has two networks, a generator and a discriminator. The generator network is trained to produce samples that follow the distribution of input data from noisy latent features. On the other hand, the discriminator is trained to differentiate the generated images from the real images. At equilibrium, it is shown that the generator network is able to produce realistic samples from the original distribution. GANs have also shown to work well in a conditional setting \cite{mirza2014conditional}, where the input noise vector to the network is conditioned on extra information. It was later shown in \cite{DCGAN}, that stable GAN configurations can be produced using deep convolutional networks. Adversarial Auto-encoders proposed in \cite{AAC}, uses adversarial learning principles to provide structure in the latent space of an auto-encoder. Here, the task of the discriminator is to differentiate between the latent samples produced by the auto-encoder from the samples drawn from a pre-determined distribution. As a result, at equilibrium, the learned latent representations follow the pre-determined distribution. 

\noindent{\textbf{Deep one-class recognition}}. One-class detection problems do not have the luxury of having the knowledge of the negative training samples. Therefore, the majority of deep one-class  detection methods have used auto-encoder structures to learn an informative latent representation. One of the earliest one-class one-class detection works based on deep learning was presented in \cite{IPMI} where the data of the given class was used to train a GAN network. During inference, the difference between a query image and it's projection on the GAN space was used to detect out-of-class samples. In a more direct application, in \cite{cvpr2018}, GANs were used in a conditional setting to force the reconstructed images of an auto-encoder network to follow the distribution of the known class. 

Recent works in one class  detection attempt to model the latent feature space of the given class. In \cite{GPND}, it is shown that the probability distribution of the learned latent space can be approximated by the product of two marginal distributions. These two distributions are approximated using the empirical distributions to be used in one-class detection.  On the other hand, in \cite{AND}, the latent representations of the given class are modeled using an auto-Regressive (AR) model.


DOCC \cite{dsvdd}, claims to be motivated from SVDD. However, different from SVDD, given the mean vector of a mini-batch, \cite{dsvdd} learns a network where the variance of data is minimized.  \textit{Hyper-sphere collapsing} is a potential problem with the method proposed in \cite{dsvdd}- where the learned network has the risk of ending up having zero weights at each layer. Despite of this limitation, \cite{dsvdd} has achieved significant improvements over the state-of-the-art methods on real image datasets.  Some of the recent deep learning-based one class classification and related methods include \cite{ocfeatures, oza2018one, baweja2020anomaly, oza2020utilizing, perera2020generative, Poojan_FG}.


\section{Proposed Method}
We begin this section with a brief overview of SVDD. Then we investigate the limitations of using auto-encoder embedding with SVDD for one-class recognition and discuss possible approaches to alleviate these limitations. We conclude this section by introducing the proposed method. 
\subsection{Overview of SVDD}
Given a set of observations $x_1, x_2, \dots x_n,$ SVDD constructs a hyper-sphere characterized by center $a$ and radii $r$ enclosing all observations. To account for possible outliers in the training set, a set of slack variables $\xi_i, i=1, 2, \dots, n$ are introduced. With this formulation, the objective can be stated as follows
\setlength{\belowdisplayskip}{0pt} \setlength{\belowdisplayshortskip}{0pt}
\setlength{\abovedisplayskip}{0pt} \setlength{\abovedisplayshortskip}{0pt}
\begin{equation}
\begin{aligned}
& \underset{a, r, \xi_i}{\text{minimize}}
& &  r^2+c\sum_{i=1}^{n} \xi_i  \\
& \text{subject to}
& & {||x_i-a||}^2 < r^2+\xi_i & \forall i, \xi_i > 0,\\
\end{aligned}
\end{equation}
where $c$ is a positive parameter that controls the trade-off between volume of the learned hyper-sphere and errors. It is shown in \cite{Tax:2004:SVD:960091.960109} that the dual of this problem can be efficiently solved using convex optimization. In our work, we denote the dual optimization objective by $l_{svdd}$. Once solved, the volume defined by the hyper-sphere characterized by center $a$ and radii $r$ is considered to be the positive space and can be used for one-class classification.

\subsection{Proposed Network}
\begin{figure}[t]
	\centering
	\includegraphics[width=0.5\linewidth]{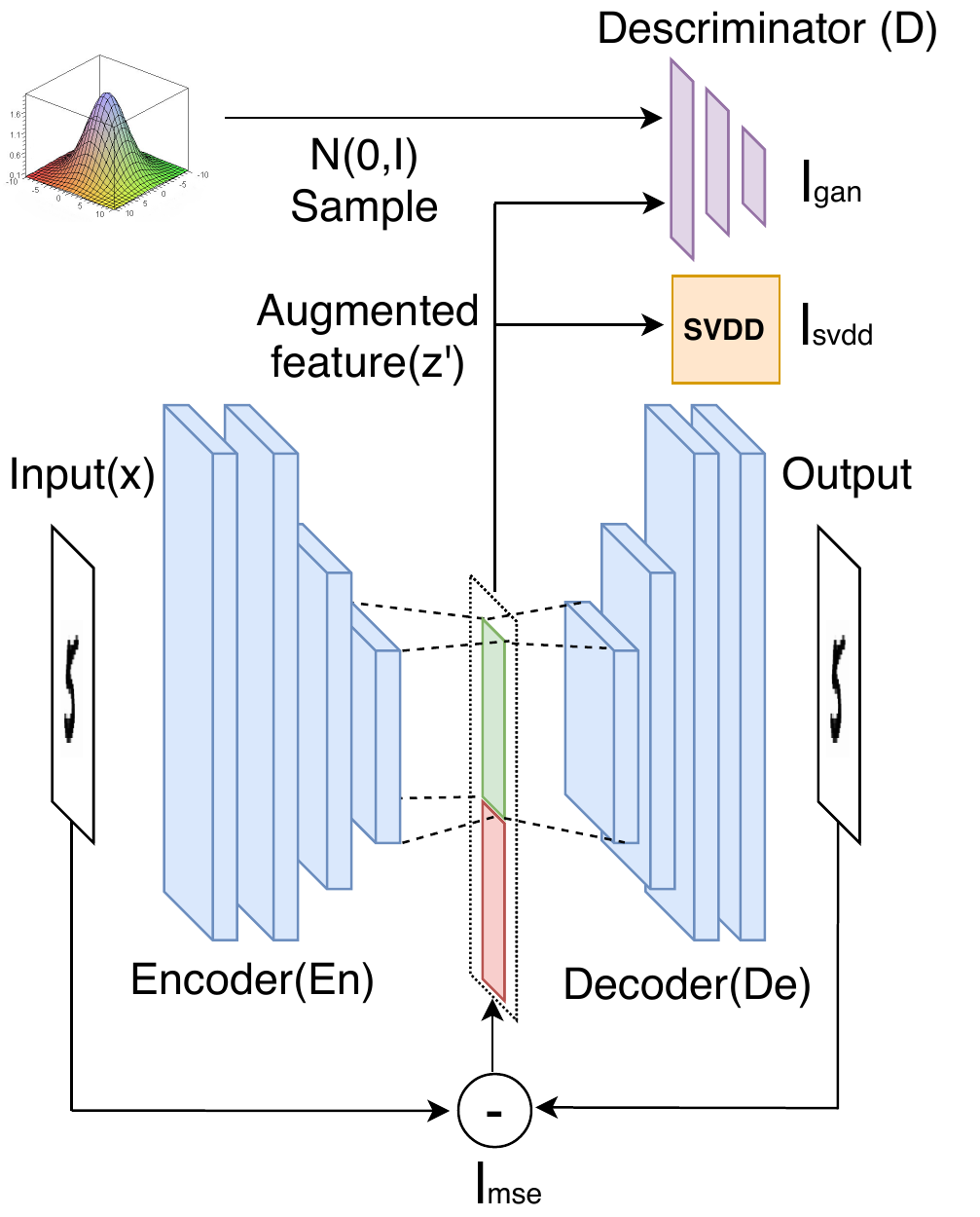}\hskip30pt
\vskip-10pt	\caption{An overview of the proposed method. We propose to use an encoder-decoder network that can transform the input observations onto a different feature space. This feature is concatenated with the reconstruction error of the network to form an augmented feature. We enforce the augmented feature to follow a specific distribution using a discriminator network to ensure that the augmented feature has SVDD friendly structure. Finally, we train an SVDD classifier on top of the augmented feature.}
	\label{fig:nw}
\end{figure}

An overview of the proposed network architecture is shown in Figure~\ref{fig:nw}. We use an auto-encoder as the foundation of our method. An auto-encoder consists of two sub networks - encoder and decoder. The encoder network $En$ is used to transform the input observations onto a new feature space $z \in R^{k}$. The obtained features are then transformed back onto the original observational space using a decoder $De$. This encoder-decoder network (collectively refereed to as an auto-encoder) is trained by minimizing the Euclidean loss $l_{mse}$ between the input and the output of the network as  $l_{mse} = \|x-\hat{x} \|^2$ where, $\hat{x} = De(En(x))$. 

With such a network architecture, the nominal region of known samples in the latent space can be identified for decision making. As mentioned earlier, this approach suffers from two notable limitations: 1) There may exist redundant space in the identified positive space. If the redundant space exists within the positive space, out-of-class objects that get projected onto such regions would result in false positives. 2) Since there exist no out-of-class observations in training, there is no way to determine the likelihood of out-of-class data appearing within the nominal region of the known class. Therefore, there are no guarantees that out-of-class samples will not get projected inside the identified positive space during inference. In the following two sub-sections, we propose ways to combat these two challenges.

\subsection{Reducing redundant space in the positive space}
\begin{figure}[t]
	\centering
	\includegraphics[width=0.45\linewidth]{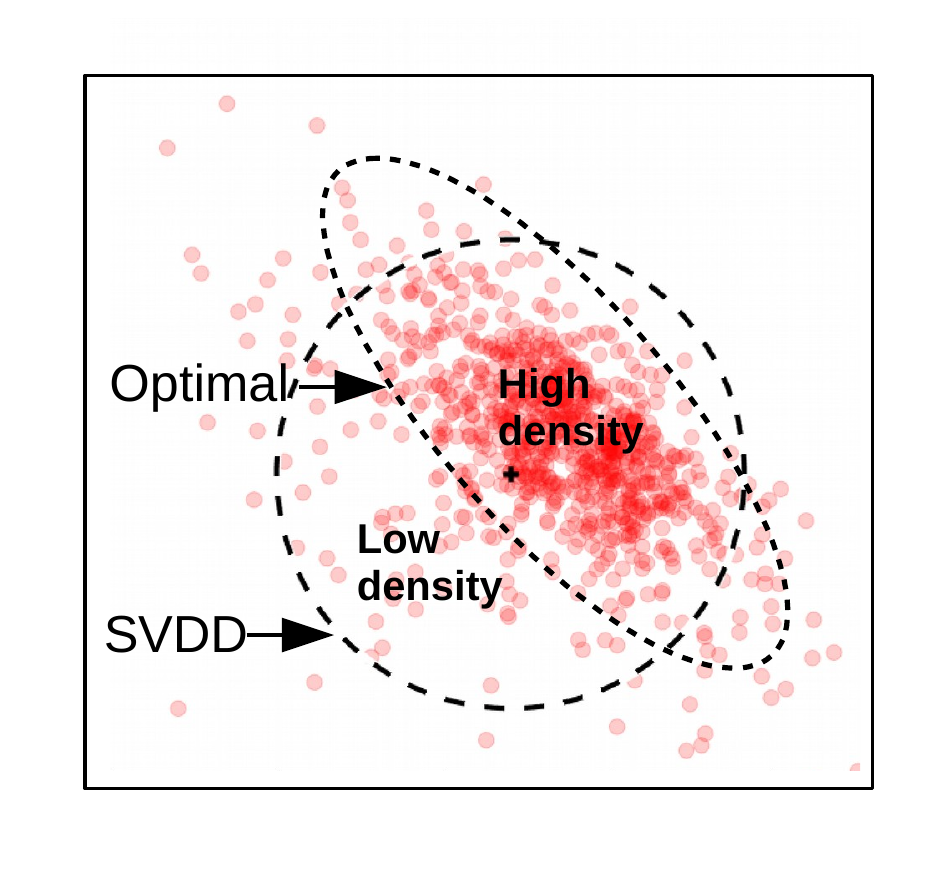} \hskip -10pt
	\includegraphics[width=0.45\linewidth]{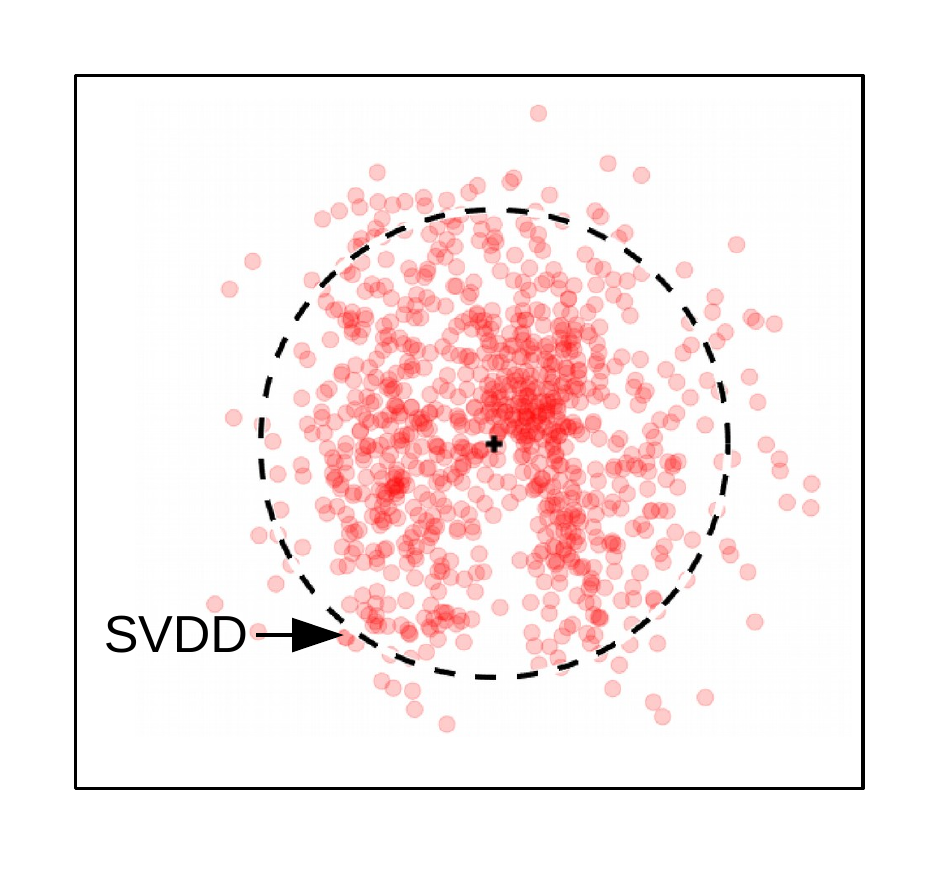}\hskip30pt
	\vskip -10pt
	\text{(a) Auto-encoder \hskip 40pt (b) Proposed.}
		\vskip -10pt
	\caption{Feature representations obtained for the MNIST digit 5 with two dimensions using (a)  Autoencoder, and (b) the proposed method. In (a), the structure of the data does not support a compact spherical boundary. When out-of-class data gets projected onto the low density region, false positive errors occur. These errors can be avoided by learning a more effective feature as shown in (b).}
	\label{fig:negative}
\end{figure}

When the SVDD formulation is devised, no assumptions are made about the underlying feature distribution. Nevertheless, SVDD is not equally effective for all types of feature distributions. Example shown in Figure~\ref{fig:negative} is a case where the hyper-spherical boundary produced by SVDD is found to be suboptimal. In order to produce a classifier with little redundant space, the underlying feature  distribution should have the following characteristics.
\begin{enumerate}
	\item {\emph{Distribution should be uni-modal.}} If multiple modes exists in the distribution, region between two modes may be sparsly populated thereby resulting in redundant space.
	\item {\emph{Distribution should be isotropic.}} This ensures that for each additional increment in radius $\Delta r$, large number of inlier points get covered in the feature space. 
	\item {\emph{Distribution should not have long tails.}} If there are long tails, SVDD will require a larger radius where the data points appear sparsely towards the tail region. As a result, redundant space will be present when the boundary is defined.	
\end{enumerate}  

In our work, we use the adversarial auto-encoder framework \cite{AAC} to enforce the latent space to have the desired structure with the aim of reducing the redundant space in the positive space. There exists a set of distributions that satisfy these properties; uncorrelated Gaussian distribution, student-t distribution and Cauchy distribution are a few examples from this set. Since the Gaussian distribution is more efficient to sample from, considering efficiency in implementation, we choose the distribution of the learned features to follow an uncorrelated Gaussian distribution. As shown in Figure~\ref{fig:nw}, our network has a discriminator network $D$. During training, the encoder network and the discriminator network are trained using the adversarial principles where the objective is to minimize the following loss 
$$  l_{gan}  =  \mathbb{E}_{s \sim N(0,I)\in \mathbb{R}^{2k} } [ log D(s) ] + \mathbb{E}_{x \sim p_{z'}}[log (1-D( z'))],$$
where $p_{z'}$ is the distribution of the augmented features. Here, the encoder network tries to produce the augmented latent features that look similar to the Gaussian $N(0,I) \in \mathbb{R}^{2k}$ vectors. On the other hand, the discriminator network tries to differentiate the augmented latent samples generated by the encoder network from the Gaussian $N(0,I) \in \mathbb{R}^{2k}$ samples. At equilibrium, the distribution of the augmented latent features are expected to roughly follow a Gaussian $N(0,I) \in \mathbb{R}^{2k}$ distribution. Therefore, based on the discussion above, the learned feature will possess properties that are helpful to train an effective SVDD classifier. 

To illustrate the effectiveness of this approach, we train two autoencoders with and without the proposed modification with a latent dimensionality of two for the MNIST dataset. Here, we consider digit 5 which has the worst performance among all digits for a conventional autoencoder coupled with the SVDD classifier. In Figure~\ref{fig:negative} we visualize the decision boundaries learned by SVDD for digit 5 in both cases and the projection of training data onto the respective latent spaces. As evident from the figure, natural structure of the data makes a hyper-spherical decision boundary sub-optimal in this case thereby resulting redundant space in the positive decision space. On the other hand, the proposed method has produced the feature space where in-class samples are densely populated in the space identified as positive by the classifier.

\subsection{Augmented Feature Space}
Next, we consider the issue of out-of-class samples getting projected onto the positive space. First, in order to illustrate this point, we revisit Figure~\ref{fig:issues}.  There, we have trained an auto-encoder network with a two-dimensional latent space for digit 1 in the MNIST dataset. In Figure~\ref{fig:issues}(b), the distribution of in-class samples in the latent space is shown along with the learned SVDD decision boundary (we set the parameter $c = 0.1$). In Figure~\ref{fig:issues}(c), we illustrate the distribution of out-of-class samples in the same space. As evident from Figure~\ref{fig:issues}(c), large number of out-of-class samples appear inside the learned SVDD boundary. In this case, if SVDD was used as a one-class classifier, these data points will account for false positive errors. In practice, when a larger latent dimension is used, a better separation between the two classes can be expected. Nevertheless, this phenomena will be present even at higher dimensions.

It should be noted that, once learning is completed, both the encoder and decoder networks represent deterministic operations. Let us assume that, once trained, for a latent sample appearing in the nominal positive space, the decoder is likely to produce an image from the known class. Therefore, if the latent representation of a out-of-class sample lies within the nominal positive space, it is likely to produce a poor reconstruction. As a result, the reconstruction error of such a sample should be high. 

To illustrate this point, we plot the mean squared error between the reconstructed images of each projected sample of Figure~\ref{fig:issues}(b) and the corresponding input images in Figure~\ref{fig:issues}(c). As evident from the figure, compared to the samples that lie outside the decision boundary, out-of-class samples that are projected inside the decision boundary have higher reconstruction errors in general. On the other hand, out-of-class samples inside the decision boundary have higher reconstruction errors compared to the known samples from the same space as shown in Figure~\ref{fig:issues}(d).

With this background, we conclude that the joint distribution of the latent feature and the reconstruction error holds more discriminative information about the novelty of a given sample. Therefore, we propose to generate an augmented feature as shown in Figure~\ref{fig:nw}. Specifically, given an input sample $x$, the reconstruction error with respect to the reconstructed sample $\|x-\hat{(x)\|^2 }$ is first evaluated. In order to give equal importance to this measure, we append this $k$ times to the latent feature to generate a $\mathbb{R}^{2k}$ dimensional augmented feature.   The augmented feature $z' \in \mathbb{R}^{2k}$ is defined as: 
$$
z'(i) =
\begin{cases}
z(i); \text{for } i\leq k\\
||x-\hat{x}||^2  ; \text{for } k \leq i < 2k,
\end{cases}
$$
where $z(i)$ is the i\textsuperscript{th} index in vector $z$. In general, the scale of the original feature $z$ and the reconstruction error $\|x-\hat{x}\|^2 $ could be different. However, scale of both quantities are brought on to a common level as a result of the adversarial procedure introduced in the following section.

\subsection{Training and Testing Procedure}
Training process of the proposed method has two steps. In the first step, the autoencoder and discriminator networks shown in Figure~\ref{fig:nw} are trained jointly using the composite loss $\lambda l_{mse}+l_{gan}$, where $\lambda$ is a constant. First, a data batch is passed through the network to obtain the corresponding latent features and the reconstruction errors. Then, the augmented feature is formed by concatenating the latent features with the reconstruction errors. Finally, the discriminator is trained using the augmented feature batch and a batch of random vectors sampled from $ N(0,I)\in \mathbb{R}^{2k} $ distribution. After the first step of training has concluded, an SVDD classifier is trained on the collection of augmented features of the given dataset. During inference, a query image is passed through the encoder-decoder structure to obtain the latent feature and the corresponding reconstruction error. Then, these elements are used together to form the augmented feature. Finally, the augmented feature is passed through the learned SVDD classifier to classify the query image.

\subsection{Network Architecture}
In our framework, we normalize and resized the input images to the size of $32 \times 32$ as a preprocessing step. The proposed encoder and decoder networks are motivated by the network architectures presented in \cite{isola2016image}.  The encoder network has a single $7 \times 7$ convolutional layer followed by three $3 \times 3$ layers. All layers have a stride of 2 and 64 channels. The decoder network is symmetric to the encoder network. A ReLU activation followed by the batch normalization was introduced after each convolution layer in the encoder network. In the decoder network, we used LeakyRelu(0.2) followed by batch norm instead. The discriminator network is a fully-connected network with linear layers with 256 and 64 neurons. 

We selected training hyper-parameters that supported the adversarial learning process. Specifically, the generator loss  is expected to decrease initially and to increase later on to reach a steady state. On the other hand, the discriminator loss is expected to decrease and reach a steady state. A base learning rate of 0.0002 was used for training. Experimentally we found that setting $\lambda = 1$, and setting the learning rate of the discriminator to be 0.01 times that of the generator learning rate yielded expected progression in adversarial learning.

\section{Experimental Results}
In this section, first we define the testing protocol and the baseline comparison methods followed by the details of datasets used in our experiments. Then, we present the performance comparison between the proposed method with the baseline methods. We conclude the section with an ablation study.

\subsection{Protocol and Performance Metrics}
We use the standard protocol and performance metric used in one-class detection in our experiments \cite{AND},\cite{dsvdd}. In one-class detection, it is common to use data from only the given class during training. In all datasets used for experiments, there exists a standard training and testing split provided. For training, we consider one class at a time and train the model only on the training data of the considered class. Then, we test on the entire test set treating all other classes as unknown classes. In our work, we repeat this process ten times for each class and report the average performance metric. As is common in one-class detection, Area Under the Receiver operating Characteristics curve (AUC-ROC) is used the measure the performance of different methods \cite{dsvdd}. Following previous works, we use the same performance metric to compare the performance of our work with the baseline methods. 

\subsection{Baseline Methods}
We compare the performance of our method with that of the following baseline methods. AE+SVDD baseline was implemented by the authors. Results of other baseline comparisons were extracted from the tables in \cite{AND}(for AND) and \cite{dsvdd}. 

\noindent \textbf{SVDD, KDE, IF}: One-class SVM, Kernel Density Estimation and Isolation Forest classifiers performed on whitened PCA features.\\
\noindent \textbf{ DCAE}: A symmetric autoencoder network where the reconstruction error is used as a one-class detector.\\
\noindent \textbf{AnoGAN \cite{IPMI}}: GAN is trained on the given class. Reconstruction error between the query and it's projection to the GAN space is used for one-class detection.\\
\noindent \textbf{SDOCC and DOCC \cite{dsvdd}}: Soft boundary version and the hard boundary version of deep one-class classification.\\
\noindent \textbf{AND \cite{AND}}: Latent feature is modeled using an autoregressive model and is used for one-class detection.\\
\noindent \textbf{AE+SVDD }: SVDD trained on auto-encoder features.\\

\subsection{Results}
\begin{figure*}[!]
	\centering
	\begin{minipage}{1\textwidth}
		\centering
		\includegraphics[width=0.7cm,height=0.7cm]{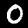}
		\includegraphics[width=0.7cm,height=0.7cm]{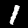}
		\includegraphics[width=0.7cm,height=0.7cm]{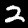}
		\includegraphics[width=0.7cm,height=0.7cm]{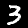}
		\includegraphics[width=0.7cm,height=0.7cm]{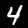}
		\includegraphics[width=0.7cm,height=0.7cm]{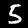}
		\includegraphics[width=0.7cm,height=0.7cm]{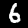}
		\includegraphics[width=0.7cm,height=0.7cm]{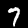}
		\includegraphics[width=0.7cm,height=0.7cm]{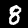}
		\includegraphics[width=0.7cm,height=0.7cm]{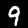}
		\includegraphics[width=0.7cm,height=0.7cm]{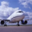}
		\includegraphics[width=0.7cm,height=0.7cm]{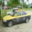}
		\includegraphics[width=0.7cm,height=0.7cm]{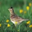}
		\includegraphics[width=0.7cm,height=0.7cm]{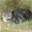}
		\includegraphics[width=0.7cm,height=0.7cm]{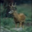}
		\includegraphics[width=0.7cm,height=0.7cm]{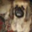}
		\includegraphics[width=0.7cm,height=0.7cm]{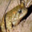}
		\includegraphics[width=0.7cm,height=0.7cm]{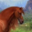}
		\includegraphics[width=0.7cm,height=0.7cm]{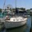}
		\includegraphics[width=0.7cm,height=0.7cm]{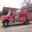}
		\includegraphics[width=0.7cm,height=0.7cm]{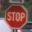}
		
	\end{minipage} 
	\begin{minipage}{1\textwidth}
		\centering
		\includegraphics[width=0.7cm,height=0.7cm]{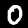}
		\includegraphics[width=0.7cm,height=0.7cm]{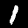}
		\includegraphics[width=0.7cm,height=0.7cm]{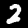}
		\includegraphics[width=0.7cm,height=0.7cm]{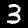}
		\includegraphics[width=0.7cm,height=0.7cm]{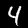}
		\includegraphics[width=0.7cm,height=0.7cm]{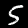}
		\includegraphics[width=0.7cm,height=0.7cm]{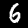}
		\includegraphics[width=0.7cm,height=0.7cm]{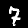}
		\includegraphics[width=0.7cm,height=0.7cm]{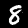}
		\includegraphics[width=0.7cm,height=0.7cm]{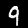}
		\includegraphics[width=0.7cm,height=0.7cm]{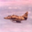}
		\includegraphics[width=0.7cm,height=0.7cm]{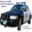}
		\includegraphics[width=0.7cm,height=0.7cm]{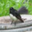}
		\includegraphics[width=0.7cm,height=0.7cm]{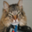}
		\includegraphics[width=0.7cm,height=0.7cm]{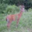}
		\includegraphics[width=0.7cm,height=0.7cm]{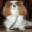}
		\includegraphics[width=0.7cm,height=0.7cm]{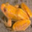}
		\includegraphics[width=0.7cm,height=0.7cm]{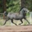}
		\includegraphics[width=0.7cm,height=0.7cm]{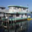}
		\includegraphics[width=0.7cm,height=0.7cm]{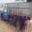}
		\includegraphics[width=0.7cm,height=0.7cm]{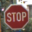}
		
	\end{minipage} 
	\begin{minipage}{1\textwidth}
		\centering
		\includegraphics[width=0.7cm,height=0.7cm]{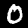}
		\includegraphics[width=0.7cm,height=0.7cm]{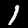}
		\includegraphics[width=0.7cm,height=0.7cm]{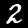}
		\includegraphics[width=0.7cm,height=0.7cm]{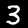}
		\includegraphics[width=0.7cm,height=0.7cm]{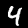}
		\includegraphics[width=0.7cm,height=0.7cm]{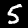}
		\includegraphics[width=0.7cm,height=0.7cm]{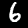}
		\includegraphics[width=0.7cm,height=0.7cm]{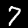}
		\includegraphics[width=0.7cm,height=0.7cm]{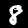}
		\includegraphics[width=0.7cm,height=0.7cm]{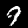}
		\includegraphics[width=0.7cm,height=0.7cm]{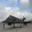}
		\includegraphics[width=0.7cm,height=0.7cm]{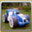}
		\includegraphics[width=0.7cm,height=0.7cm]{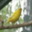}
		\includegraphics[width=0.7cm,height=0.7cm]{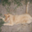}
		\includegraphics[width=0.7cm,height=0.7cm]{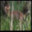}
		\includegraphics[width=0.7cm,height=0.7cm]{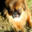}
		\includegraphics[width=0.7cm,height=0.7cm]{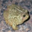}
		\includegraphics[width=0.7cm,height=0.7cm]{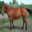}
		\includegraphics[width=0.7cm,height=0.7cm]{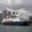}
		\includegraphics[width=0.7cm,height=0.7cm]{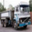}
		\includegraphics[width=0.7cm,height=0.7cm]{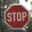}

	\end{minipage} 
	
	\vskip 8pt
	
	\begin{minipage}{1\textwidth}
		\centering
		\includegraphics[width=0.7cm,height=0.7cm]{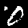}
		\includegraphics[width=0.7cm,height=0.7cm]{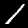}
		\includegraphics[width=0.7cm,height=0.7cm]{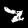}
		\includegraphics[width=0.7cm,height=0.7cm]{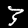}
		\includegraphics[width=0.7cm,height=0.7cm]{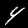}
		\includegraphics[width=0.7cm,height=0.7cm]{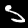}
		\includegraphics[width=0.7cm,height=0.7cm]{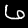}
		\includegraphics[width=0.7cm,height=0.7cm]{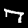}
		\includegraphics[width=0.7cm,height=0.7cm]{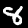}
		\includegraphics[width=0.7cm,height=0.7cm]{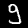}
		\includegraphics[width=0.7cm,height=0.7cm]{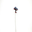}
		\includegraphics[width=0.7cm,height=0.7cm]{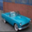}
		\includegraphics[width=0.7cm,height=0.7cm]{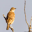}
		\includegraphics[width=0.7cm,height=0.7cm]{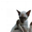}
		\includegraphics[width=0.7cm,height=0.7cm]{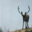}
		\includegraphics[width=0.7cm,height=0.7cm]{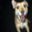}
		\includegraphics[width=0.7cm,height=0.7cm]{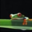}
		\includegraphics[width=0.7cm,height=0.7cm]{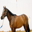}
		\includegraphics[width=0.7cm,height=0.7cm]{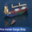}
		\includegraphics[width=0.7cm,height=0.7cm]{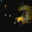}
		\includegraphics[width=0.7cm,height=0.7cm]{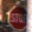}
		
	\end{minipage} 
	
	\begin{minipage}{1\textwidth}
		\centering
		\includegraphics[width=0.7cm,height=0.7cm]{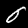}
		\includegraphics[width=0.7cm,height=0.7cm]{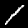}
		\includegraphics[width=0.7cm,height=0.7cm]{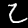}
		\includegraphics[width=0.7cm,height=0.7cm]{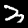}
		\includegraphics[width=0.7cm,height=0.7cm]{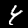}
		\includegraphics[width=0.7cm,height=0.7cm]{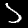}
		\includegraphics[width=0.7cm,height=0.7cm]{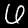}
		\includegraphics[width=0.7cm,height=0.7cm]{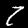}
		\includegraphics[width=0.7cm,height=0.7cm]{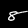}
		\includegraphics[width=0.7cm,height=0.7cm]{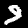}
		\includegraphics[width=0.7cm,height=0.7cm]{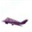}
		\includegraphics[width=0.7cm,height=0.7cm]{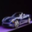}
		\includegraphics[width=0.7cm,height=0.7cm]{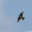}
		\includegraphics[width=0.7cm,height=0.7cm]{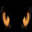}
		\includegraphics[width=0.7cm,height=0.7cm]{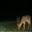}
		\includegraphics[width=0.7cm,height=0.7cm]{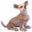}
		\includegraphics[width=0.7cm,height=0.7cm]{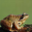}
		\includegraphics[width=0.7cm,height=0.7cm]{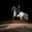}
		\includegraphics[width=0.7cm,height=0.7cm]{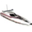}
		\includegraphics[width=0.7cm,height=0.7cm]{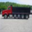}
		\includegraphics[width=0.7cm,height=0.7cm]{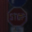}
		
	\end{minipage} 
	
	\begin{minipage}{1\textwidth}
		\centering
		\includegraphics[width=0.7cm,height=0.7cm]{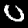}
		\includegraphics[width=0.7cm,height=0.7cm]{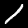}
		\includegraphics[width=0.7cm,height=0.7cm]{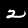}
		\includegraphics[width=0.7cm,height=0.7cm]{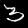}
		\includegraphics[width=0.7cm,height=0.7cm]{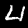}
		\includegraphics[width=0.7cm,height=0.7cm]{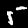}
		\includegraphics[width=0.7cm,height=0.7cm]{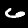}
		\includegraphics[width=0.7cm,height=0.7cm]{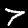}
		\includegraphics[width=0.7cm,height=0.7cm]{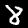}
		\includegraphics[width=0.7cm,height=0.7cm]{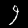}
		\includegraphics[width=0.7cm,height=0.7cm]{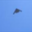}
		\includegraphics[width=0.7cm,height=0.7cm]{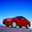}
		\includegraphics[width=0.7cm,height=0.7cm]{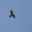}
		\includegraphics[width=0.7cm,height=0.7cm]{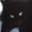}
		\includegraphics[width=0.7cm,height=0.7cm]{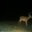}
		\includegraphics[width=0.7cm,height=0.7cm]{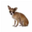}
		\includegraphics[width=0.7cm,height=0.7cm]{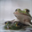}
		\includegraphics[width=0.7cm,height=0.7cm]{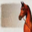}
		\includegraphics[width=0.7cm,height=0.7cm]{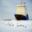}
		\includegraphics[width=0.7cm,height=0.7cm]{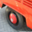}
		\includegraphics[width=0.7cm,height=0.7cm]{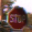}
		
	\end{minipage} 
	\vskip 8pt
	
	\begin{minipage}{1\textwidth}
		\centering
		\includegraphics[width=0.7cm,height=0.7cm]{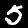}
		\includegraphics[width=0.7cm,height=0.7cm]{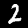}
		\includegraphics[width=0.7cm,height=0.7cm]{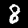}
		\includegraphics[width=0.7cm,height=0.7cm]{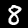}
		\includegraphics[width=0.7cm,height=0.7cm]{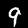}
		\includegraphics[width=0.7cm,height=0.7cm]{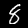}
		\includegraphics[width=0.7cm,height=0.7cm]{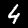}
		\includegraphics[width=0.7cm,height=0.7cm]{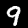}
		\includegraphics[width=0.7cm,height=0.7cm]{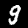}
		\includegraphics[width=0.7cm,height=0.7cm]{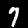}
		\includegraphics[width=0.7cm,height=0.7cm]{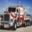}
		\includegraphics[width=0.7cm,height=0.7cm]{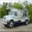}
		\includegraphics[width=0.7cm,height=0.7cm]{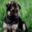}
		\includegraphics[width=0.7cm,height=0.7cm]{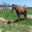}
		\includegraphics[width=0.7cm,height=0.7cm]{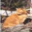}
		\includegraphics[width=0.7cm,height=0.7cm]{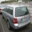}
		\includegraphics[width=0.7cm,height=0.7cm]{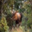}
		\includegraphics[width=0.7cm,height=0.7cm]{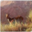}
		\includegraphics[width=0.7cm,height=0.7cm]{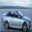}
		\includegraphics[width=0.7cm,height=0.7cm]{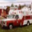}
		\includegraphics[width=0.7cm,height=0.7cm]{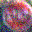}
		
	\end{minipage} 
	
	\begin{minipage}{1\textwidth}
		\centering
		\includegraphics[width=0.7cm,height=0.7cm]{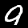}
		\includegraphics[width=0.7cm,height=0.7cm]{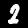}
		\includegraphics[width=0.7cm,height=0.7cm]{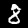}
		\includegraphics[width=0.7cm,height=0.7cm]{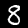}
		\includegraphics[width=0.7cm,height=0.7cm]{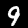}
		\includegraphics[width=0.7cm,height=0.7cm]{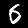}
		\includegraphics[width=0.7cm,height=0.7cm]{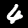}
		\includegraphics[width=0.7cm,height=0.7cm]{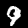}
		\includegraphics[width=0.7cm,height=0.7cm]{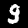}
		\includegraphics[width=0.7cm,height=0.7cm]{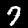}
		\includegraphics[width=0.7cm,height=0.7cm]{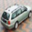}
		\includegraphics[width=0.7cm,height=0.7cm]{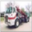}
		\includegraphics[width=0.7cm,height=0.7cm]{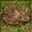}
		\includegraphics[width=0.7cm,height=0.7cm]{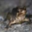}
		\includegraphics[width=0.7cm,height=0.7cm]{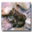}
		\includegraphics[width=0.7cm,height=0.7cm]{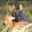}
		\includegraphics[width=0.7cm,height=0.7cm]{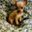}
		\includegraphics[width=0.7cm,height=0.7cm]{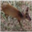}
		\includegraphics[width=0.7cm,height=0.7cm]{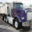}
		\includegraphics[width=0.7cm,height=0.7cm]{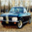}
		\includegraphics[width=0.7cm,height=0.7cm]{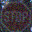}
		
	\end{minipage} 
	
	\begin{minipage}{1\textwidth}
		\centering
		\includegraphics[width=0.7cm,height=0.7cm]{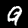}
		\includegraphics[width=0.7cm,height=0.7cm]{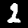}
		\includegraphics[width=0.7cm,height=0.7cm]{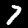}
		\includegraphics[width=0.7cm,height=0.7cm]{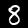}
		\includegraphics[width=0.7cm,height=0.7cm]{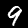}
		\includegraphics[width=0.7cm,height=0.7cm]{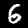}
		\includegraphics[width=0.7cm,height=0.7cm]{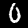}
		\includegraphics[width=0.7cm,height=0.7cm]{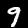}
		\includegraphics[width=0.7cm,height=0.7cm]{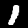}
		\includegraphics[width=0.7cm,height=0.7cm]{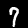}
		\includegraphics[width=0.7cm,height=0.7cm]{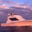}
		\includegraphics[width=0.7cm,height=0.7cm]{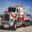}
		\includegraphics[width=0.7cm,height=0.7cm]{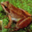}
		\includegraphics[width=0.7cm,height=0.7cm]{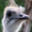}
		\includegraphics[width=0.7cm,height=0.7cm]{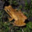}
		\includegraphics[width=0.7cm,height=0.7cm]{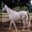}
		\includegraphics[width=0.7cm,height=0.7cm]{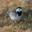}
		\includegraphics[width=0.7cm,height=0.7cm]{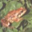}
		\includegraphics[width=0.7cm,height=0.7cm]{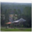}
		\includegraphics[width=0.7cm,height=0.7cm]{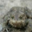}
		\includegraphics[width=0.7cm,height=0.7cm]{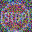}
		
	\end{minipage} 
	\vskip 5pt\caption{Qualitative performance of the proposed method. Best case predictions(top three rows), worst case predictions(middle three rows) for known classes and failed cases(last three rows) for novel classes.}\label{syn2}
\end{figure*}

Quantitative results of the GTSRB Stop sign datasets \cite{Stallkamp2012},  MNIST \cite{MNIST} and CIFAR10 \cite{CIFAR} are tabulated in Tables\ref{tblstop}, \ref{tblmnist} and \ref{cifar},  respectively. Qualitative results on the best case predictions (top three rows), worst case predictions (middle three rows) for in-class and failed cases (last three rows) for out-of-class in all considered datasets are illustrated in Figure~\ref{syn2}. Qualitative results suggests that poor predictions are made for known classes when the content of the image is significantly different from the normal images. For example, shown mis-predicted digit 6's are rotated. On the other hand, failed predictions in the out-of-class are mostly due to the high similarity in image context. For example, most mis-detections in birds class (worst performing class in CIFAR10) are with greenery images.\\

\noindent\textbf{GTSRB Stop Sign Dataset.} As a part of the evaluation of \cite{dsvdd}, the performance of one-class  detectors on the adversarial samples have been investigated. We follow the protocol defined in \cite{dsvdd} to compare the effectiveness of the proposed method in rejecting adversarial samples. In this experiment, stop sign class of the GTSRB dataset is used. The training set contains 780 images. Trained model is evaluated against adversarial attacks generated using the boundary attack. For testing, 270 normal stop sign images and 20 adversarial examples (out-of-class samples) are used. We obtained the adversarial samples from the authors of \cite{dsvdd}. Since the data size in this dataset is considerably smaller, we performed data augmentation with random horizontal flipping. In Table \ref{tblstop}, we report the mean performance obtained for our method over 50 iterations. We obtained the results of baseline  methods from \cite{dsvdd}. In this setup, SVDD operating over deep auto-encoder features obtained an AUC value of $94.0$. Comparatively, the proposed method reported an improvement of $2.6\%$ with a significantly lower variance.\\

\noindent\textbf{MNIST Dataset.} MNIST is a centered and cropped handwritten dataset of digits 0-9 with an input resolution of $28\times28$. According to Table~\ref{tblmnist}, digits 1 and 0 are the easiest two classes of the dataset with all baseline methods reporting high AUC values. In contrast, all methods have found digits 5 and 8 difficult to differentiate from the rest. On average, the proposed method obtained the best results over all ten MNIST classes.  The AND method reports the second best results along with the best results in several individual classes. However, it should be noted that unlike in other baselines, AND has only reported the performance over a single trail.\\

\noindent\textbf{CIFAR10 Dataset.} CIFAR10 is an aligned object recognition dataset with 10 object classes. Compared to the MNIST and CIFAR10 datasets, CIFAR10 contains considerable amount of intra class variation. Therefore, one-class detection performance across all methods are considerably poor compared to MNIST. Despite this fact, the proposed method is able to improve the state of the art results by nearly $6\%$ as shown in Table~\ref{cifar}. Specifically, AUC of classes frog, truck and ship have reported an AUC value of nearly $80\%$ for the proposed method. In general, all other methods have performed relatively better in the aforementioned classes as well. On the other hand, all other methods have performed poorly on the bird class and the cat class. Traditionally, one-class detection methods have not performed well on real image datasets. In this context, the performance improvement introduced in this dataset is significant.

\begin{table}[htp!]
	
	\centering
	\caption{Tabulation of average area under the ROC curve (and standard deviation) for the GTSRB Stop sign dataset.}
	\label{tblstop}
	\begin{tabular}{|l|l l|}
		\hline
		OCSVM \cite{Scholkopf:2001:ESH:1119748.1119749}& 67.5 & \textit{1.2} \\ \hline
		KDE \cite{Bishop:2006:PRM:1162264}   & 60.5 & \textit{1.7} \\ \hline
		IF \cite{Bishop:2006:PRM:1162264}   & 73.8 & \textit{0.9} \\ \hline
		DCAE \cite{MNISTAUTO} & 79.1 & \textit{3.0} \\ \hline
		SDOCC \cite{dsvdd} & 77.8 & \textit{4.9} \\ \hline
		DOCC \cite{dsvdd} & 80.3 & \textit{2.8} \\ \hline		
		Ours  & \textbf{85.2} & \textit{0.7} \\ \hline
	\end{tabular}
\end{table}

\begin{table*}[htp!]
	\centering

	\caption{Tabulation of the average area under the ROC curve for the MNIST dataset. Standard deviation values are indicated aside. *Paper had only reported the results for a single trial.}
	\label{tblmnist}
	\resizebox{\linewidth}{!}{
		\begin{tabular}{|l ||l l|l l|l l|l l|l l|l l|l l|l l|l l|l l|l l|}
	\hline
	Class & \multicolumn{2}{l|}{OCSVM\cite{Scholkopf:2001:ESH:1119748.1119749}} & \multicolumn{2}{l|}{KDE\cite{Bishop:2006:PRM:1162264}} & \multicolumn{2}{l|}{IF\cite{Bishop:2006:PRM:1162264}} & \multicolumn{2}{l|}{DCAE\cite{MNISTAUTO}} & \multicolumn{2}{l|}{ANOGAN\cite{IPMI}} & \multicolumn{2}{l|}{SDOCC\cite{dsvdd}} & \multicolumn{2}{l|}{DOCC\cite{dsvdd}} & \multicolumn{2}{l|}{AND*\cite{AND}}& \multicolumn{2}{l|}{OCGAN*\cite{Perera_2019_CVPR}} & \multicolumn{2}{l|}{AE+SVDD} & \multicolumn{2}{l|}{Ours} \\
	& & & &  & & & & &  \multicolumn{2}{l|}{(IPIM17)} & \multicolumn{2}{l|}{(ICML18)} & \multicolumn{2}{l|}{(ICML18)} & \multicolumn{2}{l|}{(CVPR19)} & \multicolumn{2}{l|}{(CVPR19)} & & & &\\ \hline \hline
			
			0     & 98.6     & \textit{0.0}    & 97.1    & \textit{0.0}   & 98.0   & \textit{0.3}   & 97.6    & \textit{0.0}    & 96.6     & \textit{1.3}     & 97.8     & \textit{0.7}    & 98.0    & \textit{0.7}    & 99.3    & \textit{0.0}  &\textbf{99.8}& \textit{0.0} & 96.8 & \textit{0.0} & {99.6}    & \textit{0.1}    \\ \hline
			
			1     & 99.5     & \textit{0.0}    & 98.9    & \textit{0.0}   & 97.3   & \textit{0.4}   & 98.3    & \textit{0.0}    & 99.2     & \textit{0.6}     & 99.6     & \textit{0.1}    & {99.7}    & \textit{0.1}    & \textbf{99.9} & \textit{0.0}    & \textbf{99.9} & \textit{0.0} & 99.3 & \textit{0.0} & 98.8    & \textit{0.7}    \\ \hline
			
			2     & 82.5     & \textit{0.1}    & 79.0    & \textit{0.0}   & 88.6   & \textit{0.5}   & 85.4    & \textit{0.0}    & 85.0     & \textit{2.9}     & 89.5     & \textit{1.2}    & 91.7    & \textit{0.8}    & 95.9 & \textit{0.0}  & 94.2  & \textit{0.0} & 83.4 & \textit{0.0}  & \textbf{97.2}    & \textit{0.5}    \\ \hline
			
			3     & 88.1     & \textit{0.0}    & 86.2    & \textit{0.0}   & 89.9   & \textit{0.4}   & 86.7    & \textit{0.0}    & 88.7     & \textit{2.1}     & 90.3     & \textit{2.1}    & 91.9    & \textit{1.5}    & \textbf{96.6}  & \textit{0.0} & 96.3   & \textit{0.0}  & 86.8 & \textit{0.0} & {95.5}    & \textit{0.3}    \\ \hline
			
			4     & 94.9     & \textit{0.0}    & 87.9    & \textit{0.0}   & 92.7   & \textit{0.6}   & 86.5    & \textit{0.0}    & 89.4     & \textit{1.3}     & 93.8     & \textit{1.5}    & 94.9    & \textit{0.8}    & {95.6}  & \textit{0.0} & \textbf{97.5}   & \textit{0.0}  & 92.4 & \textit{0.0} & 95.7    & \textit{0.4}    \\ \hline
			
			5     & 77.1     & \textit{0.0}    & 73.8    & \textit{0.0}   & 85.5   & \textit{0.8}   & 78.2    & \textit{0.0}    & 88.3     & \textit{2.9}     & 85.8     & \textit{2.5}    & 88.5    & \textit{0.9}    & {96.4}  & \textit{0.0} & \textbf{98.0}   & \textit{0.0} & 75.8 & \textit{0.0}   & 96.3    & \textit{0.5}    \\ \hline
			
			6     & 96.5     & \textit{0.0}    & 87.6    & \textit{0.0}   & 95.6   & \textit{0.3}   & 94.6    & \textit{0.0}    & 94.7     & \textit{2.7}     & 98.0     & \textit{0.4}    & 98.3    & \textit{0.5}    & \textbf{99.4} & \textit{0.0}  & 99.1   & \textit{0.0} & 93.1 & \textit{0.0}  & 98.8    & \textit{0.3}    \\ \hline
			
			7     & 93.7     & \textit{0.0}    & 91.4    & \textit{0.0}   & 92.0   & \textit{0.4}   & 92.3    & \textit{0.0}    & 93.5     & \textit{1.8}     & 92.7     & \textit{1.4}    & 94.6    & \textit{0.9}    & {98.0}    & \textit{0.0}  &\textbf{98.1}& \textit{0.0} & 92.6 & \textit{0.0}  & 95.7    & \textit{0.3}    \\ \hline
			
			8     & 88.9     & \textit{0.0}    & 79.2    & \textit{0.0}   & 89.9   & \textit{0.4}   & 86.5    & \textit{0.0}    & 84.9     & \textit{2.1}     & 92.9     & \textit{1.4}    & 93.9    & \textit{1.6}    & 95.3    & \textit{0.0}   & 93.9& \textit{0.0}    & 88.9 & \textit{0.0}  & \textbf{95.4} & \textit{0.4}    \\ \hline
			
			9     & 93.1     & \textit{0.0}    & 88.2    & \textit{0.0}   & 93.5   & \textit{0.3}   & 90.4    & \textit{0.0}    & 92.4     & \textit{1.1}     & 94.9     & \textit{0.6}    & 96.5    & \textit{0.3}    & \textbf{98.1}    & \textit{0.0} & \textbf{98.1} & \textit{0.0} & 93.7 & \textit{0.0}  & 97.7    & \textit{0.2}    \\ \hline \hline
			
			Mean  & 91.3     & \textit{0.0}             & 86.9    & \textit{0.0}            & 92.3   & \textit{0.4}            & 89.7    & \textit{0.0}               & 91.3     & \textit{1.9}              & 93.5     & \textit{1.2}             & 94.8    & \textit{0.8}             & \textbf{97.5}    & \textit{0.0}    & \textbf{97.5} & \textit{0.0} &90.2 & \textit{0.0}        & {97.1}    & \textit{0.4}             \\ \hline
	\end{tabular}}
\end{table*}

\begin{table*}[htp!]
	\centering

	\caption{Tabulation of the average area under the ROC curve for the CIFAR10 dataset. Standard deviation values are indicated aside. *Paper only reported the results for a single trial.}
	\label{cifar}
	\resizebox{\linewidth}{!}{
		\begin{tabular}{|l ||l l|l l|l l|l l|l l|l l|l l|l l|l l|l l|l l|}
			\hline
			Class & \multicolumn{2}{l|}{OCSVM\cite{Scholkopf:2001:ESH:1119748.1119749}} & \multicolumn{2}{l|}{KDE\cite{Bishop:2006:PRM:1162264}} & \multicolumn{2}{l|}{IF\cite{Bishop:2006:PRM:1162264}} & \multicolumn{2}{l|}{DCAE\cite{MNISTAUTO}} & \multicolumn{2}{l|}{ANOGAN\cite{IPMI}} & \multicolumn{2}{l|}{SDOCC\cite{dsvdd}} & \multicolumn{2}{l|}{DOCC\cite{dsvdd}} & \multicolumn{2}{l|}{AND*\cite{AND}}& \multicolumn{2}{l|}{OCGAN*\cite{Perera_2019_CVPR}} & \multicolumn{2}{l|}{AE+SVDD} & \multicolumn{2}{l|}{Ours} \\
			& & & &  & & & & &  \multicolumn{2}{l|}{(IPIM17)} & \multicolumn{2}{l|}{(ICML18)} & \multicolumn{2}{l|}{(ICML18)} & \multicolumn{2}{l|}{(CVPR19)} & \multicolumn{2}{l|}{(CVPR19)} & & & & \\ \hline \hline
			Plane & 61.6     & \textit{0.9}    & 61.2    & \textit{0.0}   & 60.1   & \textit{0.7}   & 59.1    & \textit{5.1}    &{ 67.1  }   & \textit{2.5}     & 61.7     & \textit{4.2}    & 61.7    & \textit{4.1}    & 73.5    & \textit{0.0} & \textbf{75.7} & \textit{0.0}   &  55.2  & \textit{0.0}   & 66.4    & \textit{1.5}    \\ \hline
			
			Car   & 63.8     & \textit{0.6}    & 64.0    & \textit{0.0}   & 50.8   & \textit{0.6}   & 57.4    & \textit{2.9}    & 54.7     & \textit{3.4}     & 64.8     & \textit{1.4}    & 65.9    & \textit{2.1}    & 58.0    & \textit{0.0} &53.1 & \textit{0.0}   &  73.0 & \textit{0.0}  & \textbf{78.5}    & \textit{0.6}    \\ \hline
			
			Bird  & 50.0     & \textit{0.5}    & 50.1    & \textit{0.0}   & 49.2   & \textit{0.4}   & 48.9    & \textit{2.4}    & 52.9     & \textit{3.0}     & 49.5     & \textit{1.4}    & 50.8    & \textit{0.8}    & \textbf{69.0}    & \textit{0.0}   & 64.0 & \textit{0.0}   &   49.1 & \textit{0.0} & 54.9    & \textit{0.6}    \\ \hline
			
			Cat   & 55.9     & \textit{1.3}    & 56.4    & \textit{0.0}   & 55.1   & \textit{0.4}   & 58.4    & \textit{1.2}    & 54.5     & \textit{1.9}     & 56.0     & \textit{1.1}    & \textbf{59.1}    & \textit{1.4}    & 54.2    & \textit{0.0}  &62.0 & \textit{0.0}   &  53.6 & \textit{0.0} & 57.3    & \textit{0.6}    \\ \hline
			
			Deer  & 66.0     & \textit{0.7}    & 66.2    & \textit{0.0}   & 49.8   & \textit{0.4}   & 54.0    & \textit{1.3}    & 65.1     & \textit{3.2}     & 59.1     & \textit{1.1}    & 60.9    & \textit{1.1}    & \textbf{76.1}    & \textit{0.0}  & 72.3& \textit{0.0}   & 61.1 & \textit{0.0} & 73.6    & \textit{0.1}    \\ \hline
			
			Dog   & 62.4     & \textit{0.8}    & 62.4    & \textit{0.0}   & 58.5   & \textit{0.4}   & 62.2    & \textit{1.8}    & 60.3     & \textit{2.6}     & 62.1     & \textit{2.4}    & \textbf{65.7}    & \textit{2.5}    & 54.6    & \textit{0.0}  &62.0& \textit{0.0}   &  60.4 & \textit{0.0}  & 63.1    & \textit{0.4}    \\ \hline
			
			Frog  & 74.7     & \textit{0.3}    & 74.9    & \textit{0.0}   & 42.9   & \textit{0.6}   & 51.2    & \textit{5.2}    & 58.5     & \textit{1.4}     & 67.8     & \textit{2.4}    & 67.7    & \textit{2.6}    & 75.1    & \textit{0.0} & 72.3 & \textit{0.0}   & 62.6 & \textit{0.0}  & \textbf{80.8}    & \textit{0.1}    \\ \hline
			
			Horse & 62.6     & \textit{0.6}    & 62.6    & \textit{0.0}   & 55.1   & \textit{0.7}   & 58.6    & \textit{2.9}    & 62.5     & \textit{0.8}     & 65.2     & \textit{1.0}    & 67.3    & \textit{0.9}    & 53.5    & \textit{0.0} & 57.5& \textit{0.0}   & 69.1 & \textit{0.0}  & \textbf{72.0}    & \textit{1.1}    \\ \hline
			
			Ship  & 74.9     & \textit{0.4}    & 75.1    & \textit{0.0}   & 74.2   & \textit{0.6}   & 76.8    & \textit{1.4}    & 75.8     & \textit{4.1}     & 75.6     & \textit{1.7}    & 75.9    & \textit{1.2}    & 71.7    & \textit{0.0}  & \textbf{82.0} & \textit{0.0}   & 74.7 & \textit{0.0}  & {80.3}    & \textit{0.6}    \\ \hline
			
			Truck & 75.9     & \textit{0.3}    & 76.0    & \textit{0.0}   & 58.9   & \textit{0.7}   & 67.3    & \textit{3.0}    & 66.5     & \textit{2.8}     & 71.0     & \textit{1.1}    & 73.1    & \textit{1.2}    & 54.8    & \textit{0.0}  & 55.4 & \textit{0.0}   & 77.8 & \textit{0.0}  & \textbf{79.9}    & \textit{1.0}    \\ \hline \hline
			
			Mean  & 64.8     & 0.6             & 64.9    & \textit{0.0}   & 55.5   & \textit{0.6}   & 59.4    & \textit{2.7}    & 61.8     & \textit{2.6}     & 63.3     & \textit{1.8}    & 64.8    & \textit{1.8}    & 64.1    & \textit{0.0}  & 65.7 & \textit{0.0}   &  63.6 & \textit{0.0}  & \textbf{70.7}    & \textit{0.7}    \\ \hline
	\end{tabular}}
\end{table*}

\subsection{Ablation Study}
In order to study the impact of each component of the network we carried out an ablation study using the CIFAR10 dataset. We consider five cases for the study. In what follows, we describe each case and present the obtained average area under the curve values.

\noindent \textbf{SVDD on autoencoder featurs (AUC = 63.6): } Only the encoder-decoder structure is considered. The network is optimized by minimizing the reconstruction error. Novelty detection is performed by fitting an SVDD classifier on the latent features.\\ 
\noindent \textbf{MSE on autoencoder featurs (AUC = 62.6):} Encoder-decoder structure identical to the preceding case. Novelty detection performed based on the reconstruction error.\\
\noindent \textbf{SVDD on autoencoder featurs + discriminator (AUC = 68.6):} A Discriminator is used to force the distribution of the latent feature to be Gaussian. Novelty detection performed based on the SVDD score. \\ 
\noindent \textbf{MSE on autoencoder featurs + discriminator (AUC = 63.4):} \\ Setup identical to the preceding case. Novelty detection is performed based on the reconstruction error.

\noindent \textbf{SVDD on autoencoder augmented featurs + discriminator (AUC = 70.6): } Proposed model. Novelty detection performed based on the SVDD classifier.\\

When only the Autoencoder network is used, both the SVDD output and the reconstruction error result in similar  detection performance, where the latter is marginally lower than the former. When a latent structure is enforced through a discriminator, the performance of SVDD increases significantly by $5\%$. However, the reconstruction-based one-class detection performance improves only by $0.8\%$. In the final case, where the reconstruction error is augmented to the latent feature, SVDD-based performance improves by $2\%$. Based on the evidence from the study, we conclude that both components of the proposed method play important roles in inducing an improvement in novelty detection performance.

\section{Conclusion}
In this work we presented a deep-learning based feature learning framework targeting SVDD classifier. We empirically showed that SVDD does not produce equally effective decision boundaries for all underlying distributions. We investigated qualities of underlying distributions that makes SVDD more effective. We ensured that the learned latent representation possesses these properties by enforcing latent space to have a desirable distribution. We further showed that decision making by collectively considering reconstruction error with the latent space model can improve performance in novelty detection. We demonstrated the effectiveness of the proposed method using three publicly available datasets. In the future, we hope to investigate how this method can be extended in the case of presence of pre-trained models.
\pagebreak

\section*{Acknowledgment}
This work was supported by the NSF grant 1801435.

	\bibliographystyle{IEEEtran}
\bibliography{egbib}

\end{document}